\RequirePackage{fix-cm}

\documentclass{svjour3}                     

\smartqed

\usepackage{graphicx}
\usepackage{amsmath}
\usepackage{subfig}
\usepackage{xcolor}

\begin{document}

\title{Facial Expressions Analysis Under Occlusions Based on Specificities of Facial Motion Propagation}

\author{Delphine Poux* \and Benjamin Allaert \and Jose Mennesson \and Nacim Ihaddadene \and Ioan Marius Bilasco \and Chaabane Djeraba}

\institute{D. Poux (corresponding author), B. Allaert, I.M. Bilasco and C. Djeraba \at
              Centre de Recherche en Informatique Signal et Automatique de Lille, Univ. Lille, \\ CNRS, Centrale Lille, UMR 9189 - CRIStAL -, F-59000 Lille, France \\
              \email{\{delphine.poux, marius.bilasco, chabane.djeraba\}@univ-lille1.fr}
              \and
           J. Mennesson \at
           IMT Lille Douai, Centre de Recherche en Informatique Signal et Automatique de Lille, \\ Univ. Lille, CNRS, Centrale Lille, UMR 9189 - CRIStAL -, F-59000 Lille, France \\
           \email{jose.mennesson@univ-lille1.fr}
           \and
           N. Ihaddadene \at
           ISEN Lille, Yncrea Hauts-de-France, France \\
           \email{nacim.ihaddadene@yncrea.fr}
}


\maketitle

\begin{abstract}
Although much progress has been made in the facial expression analysis field, facial occlusions are still challenging. The main innovation brought by this contribution consists in exploiting the specificities of facial movement propagation for recognizing expressions in presence of important occlusions. The movement induced by an expression extends beyond the movement epicenter. Thus, the movement occurring in an occluded region propagates towards neighboring visible regions. In presence of occlusions, per expression, we compute the importance of each unoccluded facial region and we construct adapted facial frameworks that boost the performance of per expression binary classifier. The output of each expression-dependant binary classifier is then aggregated and fed into a fusion process that aims constructing, per occlusion, a unique model that recognizes all the facial expressions considered. The evaluations highlight the robustness of this approach in presence of significant facial occlusions.
\end{abstract}

\section{Introduction}

Facial expression analysis is a field of research that has been extensively studied in recent years. Recognizing facial expressions provides information about the emotional state of a person. This information is essential in many applications areas such as health, security or communication. Indeed, it is possible to consider that a person with bad intentions in a public place can be detected by the fact that his behaviour is abnormal compared to other people. In this case, it is interesting to automatically notify the security officers in order to anticipate the danger. 

The majority of the approaches dealing with facial expressions are generally trained on unoccluded faces and give very good results. However, these approaches perform poorly when deployed on un-controlled data (e.g., video surveillance system), where the face can be highly occluded. Two types of approaches have been proposed to address challenges in presence of occlusions. The first approach tend to reconstruct the occluded parts of the face and simulate an ideal analysis context. The second approach consists in characterizing the face despite the facial occlusion and let classifier identify the closest expression among the training data. In all cases, the facial expression analysis remains challenging when occlusions occur. 

In this paper, we propose an innovative approach to overcome facial occlusions challenges. We assume that the facial movement induced by an expression is relatively close between individuals although the texture or facial geometry of each individual is highly different. The innovation brought by our contribution relies on the propagation properties of the facial movement. The movement induced by an expression spreads beyond the movement epicenter to neighbouring regions. Hence, if a region is occluded then it is possible to focus on the movement information that has been propagated to neighbouring regions. Specific facial frameworks (i.e., specific sets of facial regions) are constructed per expression, according to the importance of each facial region to recognize the underlying expression in presence of specific occlusions. Only the most relevant regions are selected in order to be robust to both small and large occlusions. The per-expression facial frameworks are then merged into a unique model in order to recognize globally any given facial expression in presence of a specific occlusion.

In Section \ref{scope}, we highlight the main contributions of the paper and discuss approaches used to handle facial occlusions challenges. The construction of optimized facial frameworks per expression in presence of given occlusions is explained in Section \ref{weight}. The merging of these facial frameworks is introduced in Section \ref{fusion}. In Section \ref{dataset}, we present the data used for learning and the experimental protocol used. Then, we present the performances obtained considering one expression at a time or all expressions simultaneously. In Section \ref{evalWeights}, we analyze the ability of the facial frameworks to recognize a given expression in presence of specific occlusions. In Section \ref{evalFusion}, we evaluate the generic expression recognition performance in presence of large occlusions and compare our approach to the other approaches from the literature. To conclude, we summarize the results and discuss perspectives in section \ref{conclusion}.

\section{Scope and background}
\label{scope}

This section highlights the main objectives of our contribution and provides a brief overview of existing approaches to handle facial occlusions.

\subsection{Background to overcome facial occlusions}

Among the approaches proposed to handle occlusions, two categories can be distinguished: approaches that reconstruct the occluded parts of the face in order to retrieve an ideal analysis context, and approaches that characterize the face despite the occlusions.

Among the reconstruction approaches, the in-painting approach is the most commonly used \cite{chen2017,jampour2017,li2017,yu2018}. In order to improve the reconstruction, recent works proposed to add some texture information from another unoccluded face. This unoccluded face is selected according to its similarities with the occluded one. Jampour et al. \cite{jampour2017} use a guidance face to help the reconstruction of an occluded face. These solutions have proven their effectiveness for the task of face recognition because a similar face is chosen to reconstruct the occluded one. Nevertheless, based on the texture, this solution does not seem appropriate for facial expression recognition task where similar expressions do not necessarily imply similar faces. Recently, new approaches based on deep neural networks and more specifically on generative algorithms networks have been proposed \cite{chen2017,li2017,yu2018}. These new approaches try to automatically reconstruct the hidden regions of the face thanks to a generative algorithm. However, these network architectures are large and parameter tuning process is complex. Besides, the large intra-face variation between individuals in the presence of facial expressions, the reconstruction of occluded regions remains relatively complex. 

Regarding the approaches characterizing facial expressions despite the presence of occlusions, they can be divided in two categories : sparse representation approaches and sub-regions approaches. Sparse representation approaches recognize facial expression on an occluded face by representing a test image as a linear combination of unoccluded images from a dictionary \cite{liu2014,huang2012}. This dictionary is composed of a set of unoccluded training images. Because the dictionary is composed of unoccluded data, occlusions cause errors in the linear combination. When these errors reach a threshold, they are implicitly considered as occlusions and are represented by an identity matrix which is isolated from the extracted facial features used by the classification process. These approaches have the advantage to implicitly localize occlusions. However, these approaches require large dictionaries covering variations for each expression in order to build accurate linear combinations and in order to have enough characteristics to discriminate between expressions. In the sub-regions approaches, the face is divided into different regions and each region is analyzed individually \cite{dapogny2018}. The results are then merged to recognize the expression. The advantage of these approaches is that they perfom well even in the absence of a large set of training data. However, the granularity of the subdivision of the face into local regions and its effect on performance remains an open question, particularly in the presence of important occlusions.

\subsection{Contribution}

Although much progress has been made in facial expression analysis field, facial occlusions are still challenging. Recent approaches proposed in the literature are not sufficient to properly characterize facial expressions in the presence of occlusions. In addition, the large intra-face variety of individuals in presence of facial expressions increases the complexity of the learning process.

\begin{figure}[!h]
\includegraphics[width=\columnwidth]{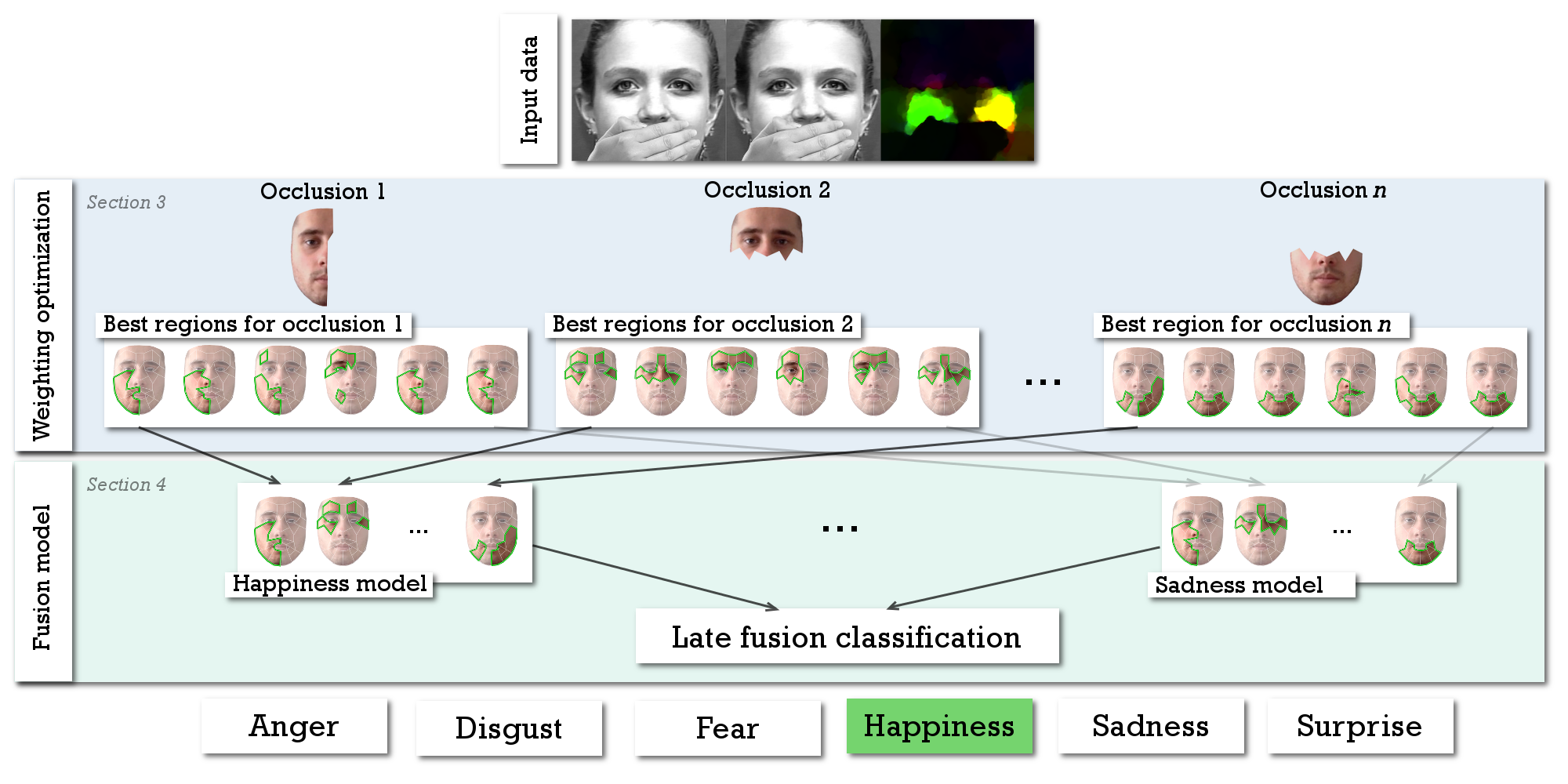}
\caption{Overview of the proposed approach.}
\label{process}
\end{figure}

Considering the descriptors used to characterize facial expressions, majority of approaches are based on texture or geometry descriptors. However, in presence of an important facial occlusion, the information to characterize the facial expression is almost completely lost or has a high probability of being noisy due to estimation errors. Recent approaches have proven the effectiveness of optical flow in characterizing facial expressions \cite{allaert2018}. Thanks to the physical properties of skin, descriptors based on movement seem adapted in the case of occlusion. Indeed, despite the fact that the epicenter of a movement is situated in an occluded part of the face, the movement related to the expression is still visible in the neighboring regions, as illustrated in Fig. \ref{process} (see input data part of the image), where the motion induced by the smile has, as a secondary effect, the rise of the cheeks.

Assuming that the movement within a face region spreads to neighbouring regions, we consider it appropriate to characterize facial expressions based on the evolution of movement through specific regions of the face. Inspired by the sub-region approaches, we propose an innovative approach to overcome facial occlusions. Fig. \ref{process} illustrates an overview of our approach which consists in recognizing facial expressions in presence of partial occlusions of the face. This approach is composed of two main steps. The first step consists in building optimized facial frameworks defining the facial regions contributing the most to the recognition of specific expressions in presence of a given occlusion. These facial frameworks are generated thanks to optimized weights computed for each facial region. These weights represent the contribution of each region to recognize a particular expression. The most important ones are selected in order to construct dedicated facial frameworks. The second step, illustrated in the lower part of Fig. \ref{process}, takes advantage of the obtained facial frameworks in order to train one binary model per expression. The results obtained with these binary models are then merged and a unique model per occlusion is trained in order to classify all expressions. 

\section{Weighting optimization algorithm}
\label{weight}

In this section, we investigate the best compromise between the minimum number of facial regions required to recognize facial expression and the performance obtained in different occlusions. 

\subsection{Weighting facial region scheme}

The weighting algorithm consists in three steps. The first step generates various partial facial frameworks (a subset of facial regions), called configurations, including fewer regions than the initial facial framework. Inspired by \cite{allaert2018}, we consider a facial framework using $25$ regions laid out following the facial muscle scheme. For each configuration $C_j$, the weighting algorithm evaluates the performance of the classification process using only the motion information contained in the regions $R_{ik}$ composing $C_j$. Then, the recognition rate obtained for a given configuration $C_j$ serves to infer the contribution of each region $R_{ik}$ to the classification process. 

\subsubsection{Configurations}

\begin{figure}[!h]
\centering
\includegraphics[width=0.5\columnwidth]{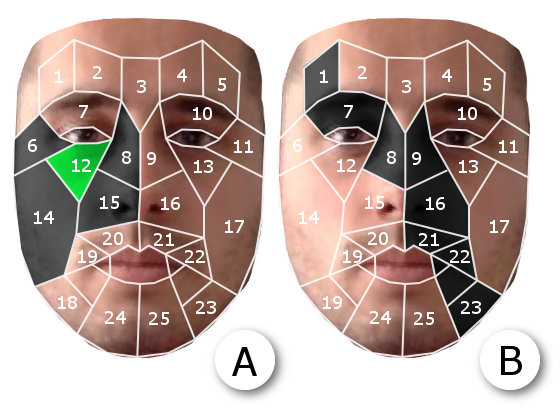}
\caption{Neighboring configurations.}
\label{facialFramework}
\end{figure}

The choice of the retained configurations in the weighting algorithm is essential. Generating the whole set of configurations that covers all combinations of one to twenty-five regions is heavy and time consuming. Instead, in order to reinforce the motion propagation properties, we decided to consider only configurations containing pair-wise connected regions.
As illustrated in Fig. \ref{facialFramework}-A, from the region $R_{12}$, the combinations $\{R_{12}\}$, $\{R_6,R_{12}\}$, $\{R_8,R_{12}\}$, $\{R_{14},R_{12}\}$ and $\{R_{15},R_{12}\}$ of size one and two are obtained. Indeed, the regions $R_6$, $R_8$, $R_{14}$ and $R_{15}$ are directly connected to the region $R_{12}$. Bigger combinations are obtained using the pair-wise connectivity of regions.

We have chosen to explore configurations containing less than $8$ regions as these configurations cover already horizontal, vertical and diagonal parts of the face as illustrated in Fig. \ref{facialFramework}-B. The configuration construction process guarantees that the configurations cover several muscles of the face and enable us to study the correlation between them. 

\subsubsection{Transferring weights to regions} 

The collected results obtained from all configurations are directly used to compute each region weight. At the beginning, each region receives a zero weight. Then, the classification rate obtained for each configuration $C_j$ is used to compute a score according to the mean classification rate of all configurations normalized by the standard deviation. This score is calculated as follows :

\begin{equation}
\omega(C_j) = exp((res_{C_j}-mean_i)/std_i)/exp(i)
\end{equation}

\noindent where $i$ is the number of regions of the configuration $C_j$, $j \in [1, 21294]$, $i = |C_j| \in [1,8]$, and $res_C$ refers to the result obtained with $C_j$. $mean_i$ and $std_i$ are respectively the mean and the standard deviation of the results obtained with all the configurations containing $i$ regions. Finally, $exp(i)$ which is the exponential of $i$, moderates the contribution of each configuration with regard to its size. Indeed, configurations covering larger portion of the face are expected to provide higher recognition rates. 

The score obtained is then added to the current weights of each region $R_{jk}$ included in the configuration $C_j$. Finally, each region weight is normalized with regard to the number of apparition of the region in all the combinations.

\begin{figure}[!htpb]
\centering
\includegraphics[width=\columnwidth]{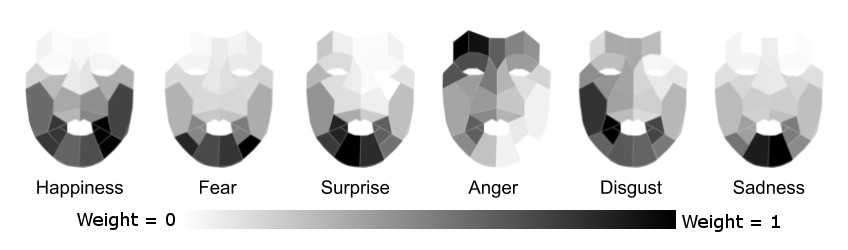}
\caption{Heatmaps of the importance of regions per facial expression.}
\label{heatmaps}
\end{figure}

The obtained weights reflect the importance of each region for recognizing each expression. Fig. \ref{heatmaps} illustrates the heatmaps obtained on CK+ \cite{lucey2010} dataset using an SVM classifier with RBF kernel and a 10-fold cross-validation protocol. This figure reveals that for almost all expressions the bottom of the face is activated, except for the anger which mainly activates the eyebrows regions. 

\begin{figure}[!h]
\centering
\includegraphics[width=\columnwidth]{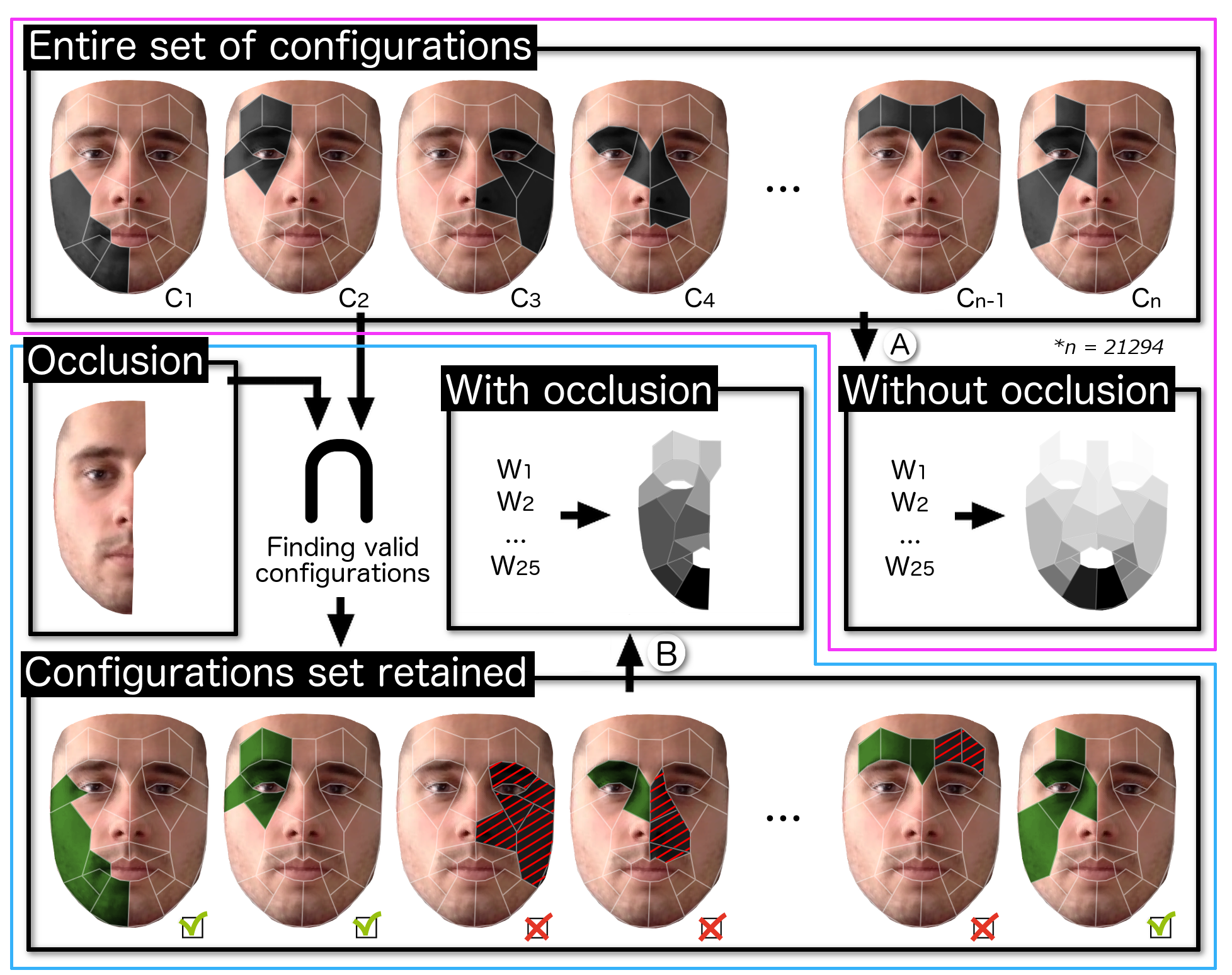}
\caption{Weights transfer considering facial occlusion (sadness expression).}
\label{weight_with_occlusion}
\end{figure}

The weight transferring process is represented in Fig. \ref{weight_with_occlusion}. The example shows the construction of the heatmap for the sadness expression. Fig. \ref{weight_with_occlusion}-A (bounded by the purple border) presents the heatmap obtained in absence of occlusions and in Fig. \ref{weight_with_occlusion}-B (bounded by the blue border) presents the heatmap obtained in presence of one occlusion. 

As seen in the weighting heatmap obtained without any occlusion (i.e. considering entire set of configurations $C$), the most important regions for this expression are situated under the mouth. Considering the weight evaluation in presence of occlusions, the process is very similar to the unoccluded situation, but during the weight transferring part, we only consider configurations $C$ that include only unoccluded regions (e.g., corresponds to the checked green configurations in Fig. \ref{weight_with_occlusion}-B). Thus, the importance of each visible region is computed independently from the occluded regions. Besides, occluded regions have a zero weight at the end of the process. This result is completely consistent because an occluded region gives no information about the facial expression.

As seen in the heatmap computed in presence of an occlusion impacting all the right part of the face, all configurations involving right regions are filtered out before transferring weights to regions. The resulting heatmap has zero weights for all regions on the right side of the face (blank areas) and the weights on the left side of the face are different with regard to the unoccluded heatmap notably for the cheek regions.

\subsection{Optimizing facial framework for expressions recognition}

\begin{figure}[!h]
\centering
\includegraphics[width=\textwidth]{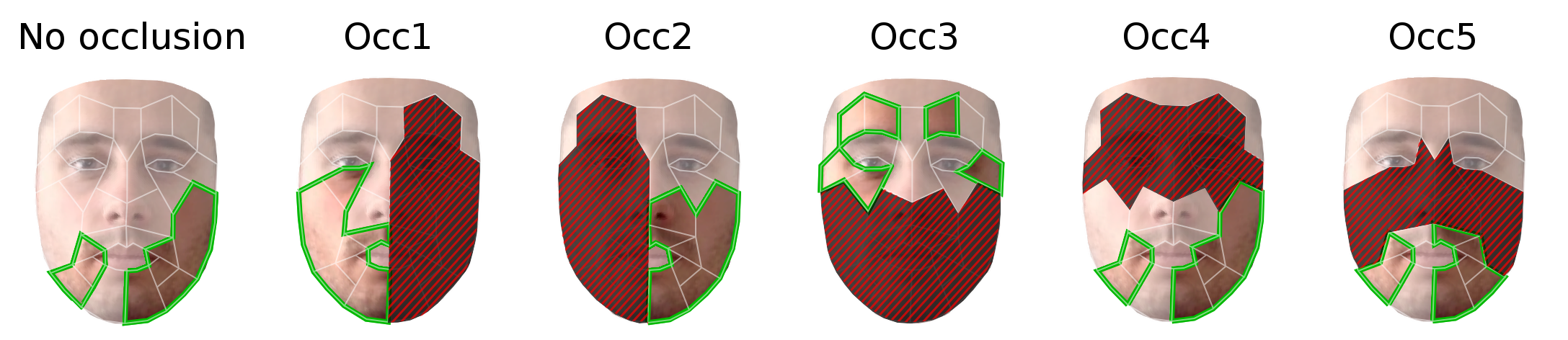}
\caption{Best facial frameworks for happiness expression under different occlusions.}
\label{happinessff}
\end{figure}

The regions are sorted according to their weights in order to determine the ranking of the regions for each expression. This ranking is then used to generate models containing from one to twenty-five regions. Each model contains the $n$ best regions for each facial expression. The obtained results reveals : a) the optimal facial frameworks for each facial expression; and b)the minimal number of regions required to recognize the expressions with performances similar to those obtained in absence of occlusion. These facial frameworks are illustrated in Fig.\ref{happinessff} for the expression of happiness in presence of different occlusion patterns by selecting the $6$ best regions.

\section{Fusion of facial expression models}
\label{fusion}

\begin{figure}[!h]
\centering
\includegraphics[width=\textwidth]{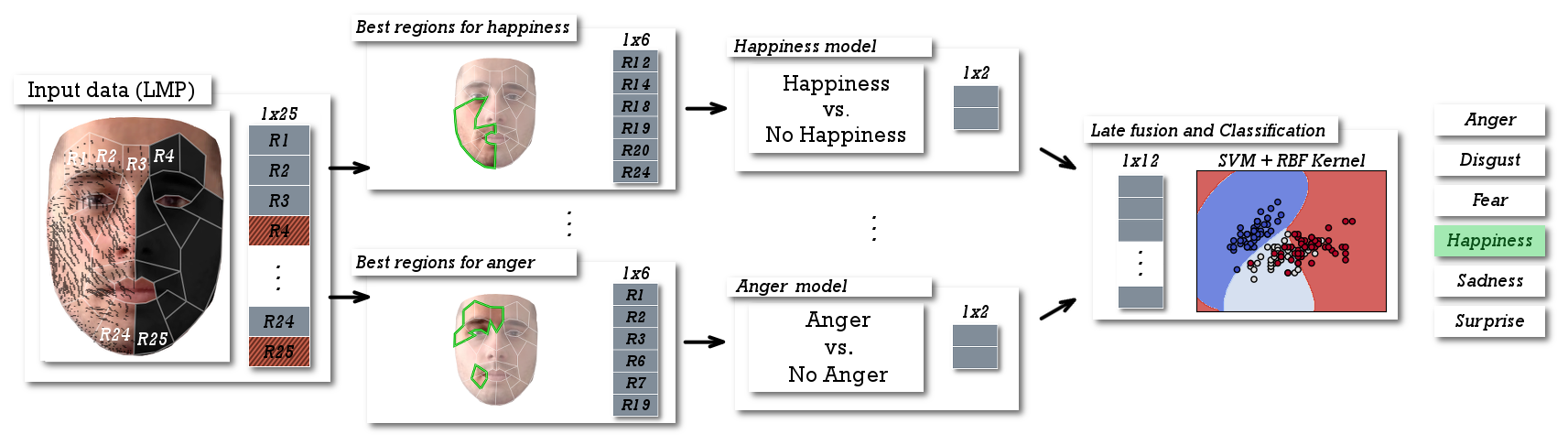}
\caption{Overview of the fusion process.}
\label{fusion_fig}
\end{figure}

The weighing optimization algorithm allows the construction of one model per expression and per occlusion. Each model corresponds to a binary classifier and indicates, per expression, if the input data corresponds to the underlying expression or not. In order to recognize an expression, regardless of the binary classifiers, we add a fusion step and, hence, construct a unified model for all expressions. The overview of the whole process are illustrated in Fig. \ref{fusion_fig}. 

As we build a learning architecture using two layers, we proceed with two learning processes : one concerning the binary classifiers and one concerning the fusion layer. For each learning process, an adapted training set is prepared.

At first, six models are trained to recognize one expression against the others. In this case, we need a training set per expression (one expression against the others). For each model, the regions are first selected according to the $x$ best regions that characterize an expression under a specific occlusion.

The constructed facial frameworks are then used to train, per expression, binary classifiers. The outputs of these models represent : a) the probability of an input sample to be classified as the underlying expression; and b) the probability of an input sample to belong to a different expression class.

A new training step is then performed with another training set which covers all expressions. In order to do this second training, raw data is fed to each binary classifier previously trained. The binary classifiers are not trained anymore and the models do not change. They are used only to compute, per expression, the probabilities that the input sample belongs or not to a specific expression class. These probabilities are concatenated into feature vectors that are fed into the fusion process. 

\section{Evaluation protocol}
\label{dataset}

In this section, we present the protocol used to conduct our evaluations. First, we introduce the descriptors used to characterize facial movements. Then, we detail the dataset and the selected facial occlusions. 

\subsection{Facial motion characterization}
\label{motion}

Recently, Allaert et al. \cite{allaert2018} proposed a descriptor called Local Motion Patterns (LMP). It characterizes the facial movement by retaining only the main directions related to facial expressions, while avoiding motion discontinuities. 

In order to characterize the movement within the face, LMPs are applied to small regions that are laid out on the face according to the facial muscles scheme. Hence, according to the relevance of the movement in the presence of facial expressions and the location of facial muscles, the face is segmented into 25 regions. For experiments, we use the same segmentation to characterize expressions when occlusions occur.

Within each facial region, LMP filters the movement computed by the optical flow using the coherence in terms of direction and intensity. Considering the elastic properties of the skin, a coherent movement gradually spreads in its neighborhood respecting, to some extend, the initial direction and intensity. Recursively, LMP analyzes the motion distribution (direction and intensity) $FDMH_{LMR_{x,y}}$ on small patches (LMRs) within each facial region. If there is a correlation between the distribution of these LMRs, then it means that there is a coherent motion propagation action taking place. This implies that the captured movement has a greater probability of reflecting a real expression and that it is not linked to a discontinuity in the movement.

When a facial region associated with an LMP is analyzed, the set of coherent LMR motion distributions are summed. Two motion distributions are coherent if there is a strong correlation in terms of main directions and intensities between them. The summed distribution results in a magnified direction histogram $MD_{LMP_{x,y}}$ calculated as follows:

\begin{equation}
\begin{split}
MD_{LMP_{x,y}}
& = \{ \sum_{i=0}^{n} FDMH_{LMR_{x,y}} \\
& \mid FDMH_{LMR_{x,y}} \in LMP_{x,y} \}.
\label{eq0}
\end{split}
\end{equation}

\noindent where $n$ represents the number of coherent LMR that composes the LMP. The intensity of each direction is then represented by the number of co-occurrences of each direction bin within the different LMRs.

The characterization of the global facial movement by LMPs consists in calculating $MD_{LMP_{x,y}}$ for the 25 facial regions. To reinforce the coherence of the movement, each region is analyzed over the entire expression sequence. Each $MD_{LMP_{x,y}}$ of the same region during the video sequence is aggregated within a space-time histogram $STMD_{LMP_{x,y}}$, as follows :

\begin{equation}
{STMD_{LMP_{x,y}}}^{k} = \sum_{t=1}^{time} {MD_{LMP_{x,y}}}_{t}^{k}.
\label{eq10}
\end{equation}

\noindent where $t$ is the frame index and $k = 1,2,...,25$ is the facial region index. Finally, histograms ${STMD_{LMP_{x,y}}}^{k}$ are concatenated into one-row vector $GMD$, which is considered as the feature vector for the global facial movement $GMD = ({STMD_{LMP_{x,y}}}^{1},{STMD_{LMP_{x,y}}}^{2},...,{STMD_{LMP_{x,y}}}^{25})$. The feature vector size is equal to the number of ROI multiplied by the number of bins. This vector is then used in our assessments to characterize facial expressions.

\subsection{Dataset}

The proposed approach is evaluated on the CK+ dataset as it is one of the most frequently dataset used in the literature to handle occlusions \cite{cornejo2018,dapogny2018,huang2012,li2018} and it contains video sequences which are adapted to study the movement. CK+ is a controlled dataset which contains 374 labelled video sequences. Each video sequence starts from the neutral face and ends with the apex of the expression.

In this dataset, images do not contain any occlusions, so, they have to be simulated. On one hand, occlusions are not totally realistic and there is a little gap between a real occlusion and a simulated one. But, on the other hand, we can totally control the experiments. By controlling the occlusion process, we can clearly quantify its impact on the overall performance. Besides, it offers also the possibility to construct precise benchmarks for comparison purpose.

\subsection{Selected facial occlusions}

\begin{figure}[!h]
\centering
\includegraphics[width=\columnwidth]{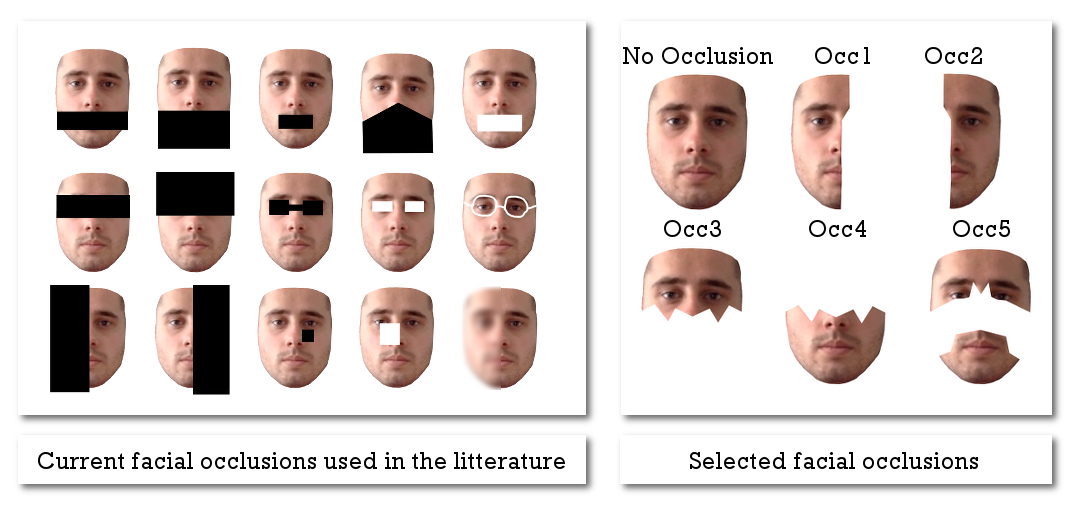}
\caption{Selected occlusions according to those used in the literature.}
\label{occlusions_fusiondataset}
\end{figure}

Generally, the occluded regions are located at the level of the mouth and eyes, under different sizes. In order to simulate head pose variation, some approach hide half of the faces (right or left). Occlusion are often generated by the altering parts of the face by adding white, black or noisy pixels. Sometimes a blur effect can be applied instead. Some examples are presented in the left part of Fig. \ref{occlusions_fusiondataset}.

Not having a stable and widely accepted baseline to compare the performance of our approach on occluded faces, we choose to simulate important occlusions to challenge our approach, as illustrated in the right part of Fig. \ref{occlusions_fusiondataset}. 

Inspired by the wide range of occlusions used in the literature, we choose a limited set of occlusions but which covers all the challenges. Indeed, the occlusion considered in our study present occlusions that impact larger facial area than those usually met in the literature.

The first two configurations (Occ1 and Occ2) present important occlusions on the left and right parts of the face. The third occlusion is inspired by the observations of Kotsia et al. \cite{kotsia2008} that underline the fact that the mouth has a great importance to recognize expression. Hence, in order to strongly challenge our approach, we define an occlusion configuration that impacts the mouth, the cheeks and the nose. Two other configurations consider important occlusions appearing on the upper part of the face and occlusions appearing in the middle part of the face.

\section{Evaluation per expression}
\label{evalWeights}

We propose a per expression evaluation in order to check if the constructed facial frameworks provide interesting results. We first evaluate the impact of the region selection when there is no occlusion. This first evaluation allows us to evaluate the accuracy of our weight calculation and, also, to find a minimal number of regions required to recognize an expression.
Then, we evaluate the efficiency of our per expression recognition method in presence of the selected occlusions. 

\subsection{Experimental protocol}
\label{cbmi_protocol}

In order to work individually with each expression and to build relevant model for per-expression recognition task, we generated several subsets of the CK+ dataset per expression. In each newly generated subset, all the sequences available for one expression are compared to a randomly stratified combination of all other expressions. For example the happiness subset contains two classes : happiness versus no-happiness. All the videos labeled happiness from the initial dataset are kept. For the no-happiness class, videos labeled with the five other expressions are randomly picked and a stratification scheme is employed in order to guarantee the same representativity of the other expressions as in the initial dataset.

For this evaluation, we have generated $25$ configurations using one region, $46$ configurations using two regions and so on until $12827$ configurations using 8 regions. A total number of $21294$ configurations are generated. The $21294$ configurations are generated for each expression and all these models are sent to SVM classifiers. Weights are calculated for the twenty-five regions and for each expression. The regions are ranked according to the computed weights. The ranking is further used to generate twenty-five models by facial expression containing from one to twenty-five regions. 

\subsection{Impact of the selection process}
\label{cbmi_results}

\begin{figure}[!h]
\includegraphics[width=\columnwidth]{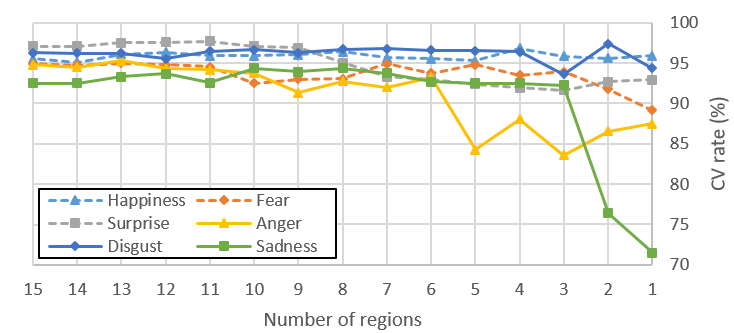}
\caption{Expression recognition rate according to the number of regions.}
\label{weightsCK_emotion}
\end{figure}

Fig. \ref{weightsCK_emotion} shows the results obtained for each facial expression using the sorted regions. These results show that this approach allows to be robust to really important occlusions. Indeed, facial expressions corresponding to surprise, happiness and disgust have quite optimal results with only one region. Sadness must have at least three regions to be recognized and anger needs at least six regions. This result is related to the complexity of the emotion. The anger and disgust expressions shares the same facial regions, which makes it hard to distinguish them with fewer regions. It is then necessary to take into consideration a larger number of regions.

\subsection{Efficiency of the approach on occluded faces}

New weights are calculated for each occlusion and each facial expression to get specific models robust to the considered occlusion. For each expression and for each occlusion, the model which gives the best result is selected. We report results on the models containing the six best regions calculated for the occlusion on the facial expression. These 6-regions models are more stable with regard to some occlusions than the full facial framework using 25 regions. They reasonably limit the number of required unoccluded regions.

\begin{figure}[!h]
\includegraphics[width=\columnwidth]{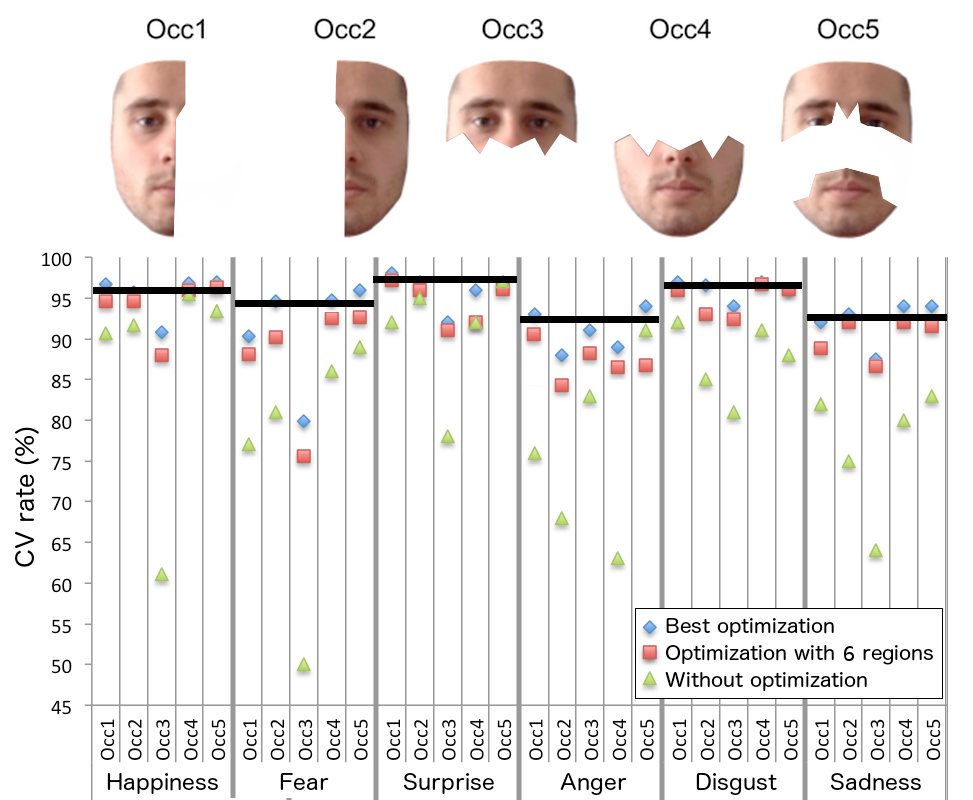}
\caption{Performance comparison with occlusion by expression on CK+.}
\label{resultsCK}
\end{figure}

Fig. \ref{resultsCK} shows the results obtained with and without intelligent facial frameworks per expression in presence of occlusions. The results obtained without intelligent facial frameworks are calculated with a model trained on unoccluded data using entire facial framework composed of the twenty-five regions. The optimized results are obtained with our approach by taking the best results considering the visible regions on one hand, and by taking the results given with the six best regions on the other. Finally, the black lines represent the results obtained in the case of an entire unoccluded dataset. According to these results, it is clear that our approach improves significantly the results even if only six regions are used. Indeed, considering the results provided by the best six regions calculated for each occlusion and for each expression gives results close to the best results obtained in unoccluded settings.

With regard to the obtained results, except for anger, the lower part of the face is really important for almost almost all expressions. Indeed, without optimization, the worst results are obtained with the mouth occlusion and the proposed approach significantly increases the performances in all situations. Concerning the anger expression, we can see that a lot of information is localized around the eyes. However, interesting results are obtained in spite of eyes occlusion.  Finally, we can see that the occlusion of the nose and cheeks have an impact more or less important according to the expression. This observation shows that the effect of propagation of the movement give non negligible information.

\section{Evaluation for all expressions}
\label{evalFusion}

In this section, We present an evaluation of the entire process. We first evaluate the effectiveness of our approach to characterize the six universal expressions (happiness, anger, disgust, fear, sadness and surprise) under different facial occlusions. Then, a comparison with representative approaches from the literature  is performed. 

\subsection{Experimental protocol}

\begin{figure}[!h]
\centering
\includegraphics[width=0.58\columnwidth]{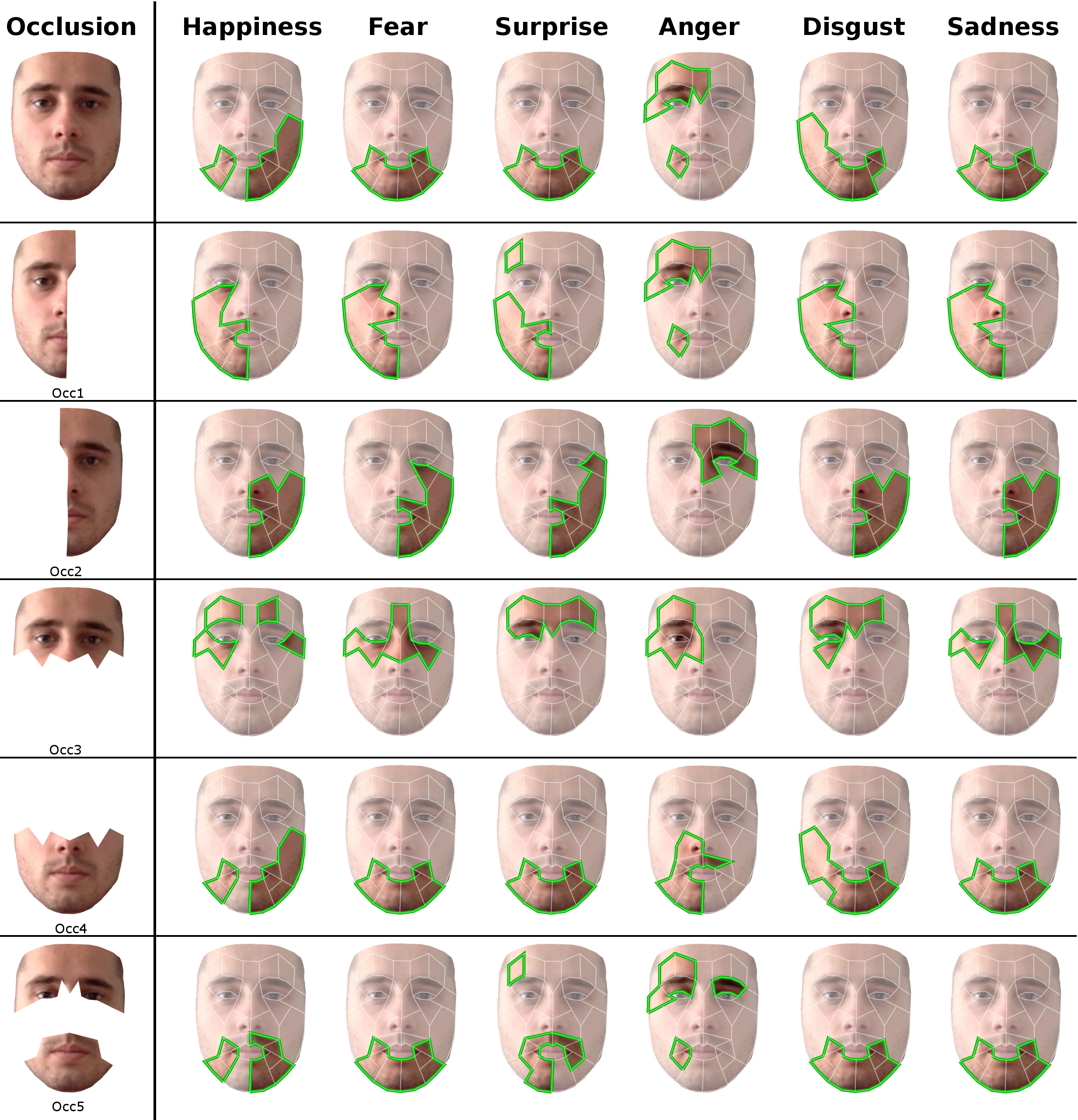}
\caption{The six best regions according to occlusions per expression.}
\label{selectRegions}
\end{figure}

In our evaluation, we selected the six best regions for each expression for each occlusion calculated. The selected facial frameworks for each expression per occlusion are illustrated in Fig. \ref{selectRegions}.

In order to evaluate our approach, we had to split the dataset in two training sets : one to train the per expression models and another to train the fusion model. We take 40\% of the sequences to build the first set for the first training sets. The remaining 60\% are then used for the second training set for the fusion. 

For the first training sets, we need six different training sets : one per expression. In order to build these training sets, we take all sequences for the current expression. In order to have balanced distribution, a same number of data for the expression and for the others expressions are respected. Thus, we randomly pick $1/5$ of the number of data of the expression for the five other expressions. By randomizing the initial dataset, we then created ten different sets of the training sets. 

These results are calculated for the ten training sets and we report the mean result obtain for the ten runs. For each training set, an SVM classifier with RBF kernel are used with a 10-fold cross-validation protocol. Then, the average classification rates are reported.

\subsection{Performances analysis}
\label{fusion_results}

In order to evaluate the performance of our approach, we study three criteria. At first, we analyze the recognition performance of facial expressions in the presence of different occlusions. Then, we compare our performance with other approaches proposed in the literature.

\subsubsection{Performances analysis under occlusions}

In this section, we analyze the performances of our approach to characterize the six universal facial expressions (happiness, sadness, disgust, fear, surprise and anger) under different occlusions. Table \ref{ourResults} shows the results obtained with our process with and without occlusion. The process without occlusion considers the six most important regions of the face to recognize each expression. The process with occlusion consider the calculated regions considering the several occlusions.

\begin{table}[!h]
`\centering
\caption{Accuracy of our approach on CK+ dataset with and without occlusion.}
\label{ourResults}       
\begin{tabular}{cccccc}
\hline\noalign{\smallskip}
No occlusion & \raisebox{-.5\height}{\includegraphics[width=0.5in]{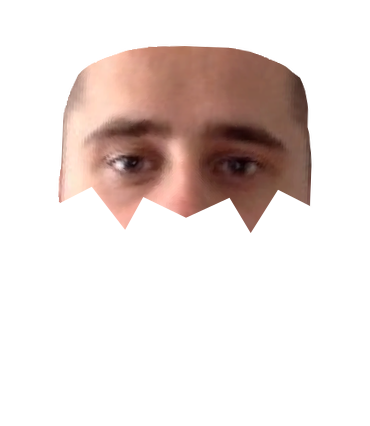}} & \raisebox{-.5\height}{\includegraphics[width=0.5in]{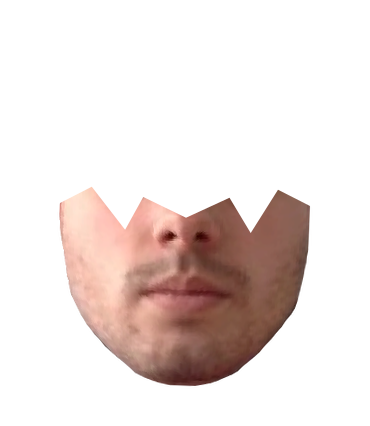}} & \raisebox{-.5\height}{\includegraphics[width=0.5in]{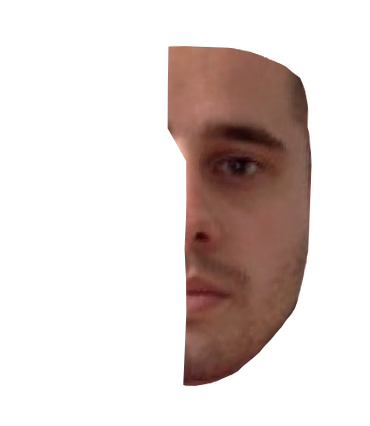}} & \raisebox{-.5\height}{\includegraphics[width=0.5in]{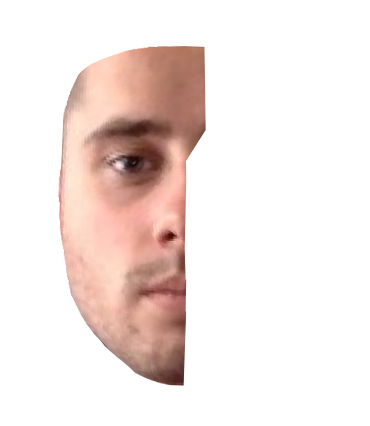}} & \raisebox{-.5\height}{\includegraphics[width=0.5in]{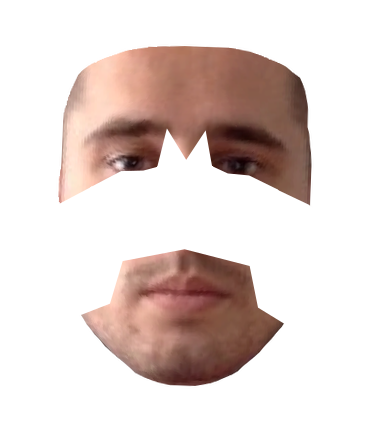}}  \\
\noalign{\smallskip}\hline\noalign{\smallskip}
91.35\% & 71.9\% & 88.8\% & 88.6\% & 88.8\% & 90.6\% \\
\noalign{\smallskip}\hline
\end{tabular}
\end{table}

As observed in Table \ref{ourResults}, we can conclude that the proposed approach is relatively robust in the presence of severe facial occlusions. As we can see, the results obtained with an occlusion of the bottom of the face drop significantly. It demonstrates the importance of the mouth regions about the expression and it is harder to compensate with the information found in the upper part of the face. Concerning the eyes, although these regions are really important for some expressions, the results obtained with our process give good results. 

Although some occlusions tend to reduce performance, the difference with the unoccluded face is not significant. This shows that the analysis of the propagation of movement is a good solution to overcome partial facial occlusions.

\subsubsection{Performances comparison with others approaches}

\begin{figure}[!htpb]
\centering
\includegraphics[width=0.7\columnwidth]{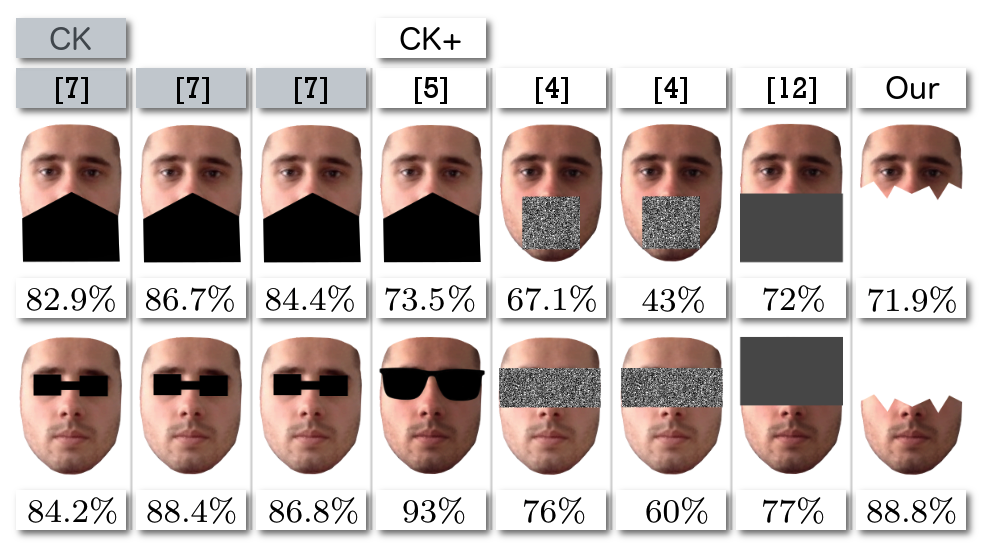}
\caption{Comparison of performances with others approaches in the litterature.}
\label{perf_results}
\end{figure}

In this section, we compare the performance of our approach with the other approaches proposed in the literature on the CK+ database. Since there is no predefined baseline to compare the different approaches, we only analyze the occlusions that are closest to the other approaches. The results are represented in Fig. \ref{perf_results}.

In view of the results, our approach gives very competitive performances. It is important to note that our occlusions are more severe than those used in other approaches except for \cite{ranzato2011} which may explain the difference with some approaches. Indeed, cheek level information for the first line and forehead level information for the second line are important to characterize facial expressions. 

\section{Conclusion}
\label{conclusion}

In this paper, we design an approach that handle expression recognition in presence of occlusions. We propose, as a first step, a method to calculate a facial framework for each expression adapted to a considered occlusion. Based on the calculated facial frameworks, we propose then a fusion step in order to build an entire process which takes an input data and predict, at the end, the expression. 

In order to do that, we pre-trained several models: one per expression in order to get the probabilities that the input data belongs to an expression class. These probabilities are then aggregated and they are used for training the fusion model.

The results obtained with this process are competitive with state-of-the-art methods, although we have considered larger occlusions. Nevertheless, it is still difficult to compare with other approaches especially due to reproductibility issues. One of our future work consists in building a benchmark regrouping a large set of occlusions and allowing the community to benefit from a stable evaluation framework.

\bibliographystyle{spmpsci}      
\bibliography{biblio}   

\end{document}